\documentclass[lettersize,journal]{IEEEtran}
\pdfoutput=1
\usepackage{amsmath, amsfonts, amssymb, amscd, mathtools, physics}
\usepackage{algorithmic}
\usepackage{array}
\usepackage[caption=false,font=normalsize,labelfont=sf,textfont=sf]{subfig}
\usepackage{textcomp}
\usepackage{stfloats}
\usepackage{url}
\usepackage{verbatim}
\usepackage{graphicx}
\usepackage{subcaption}
\def\BibTeX{{\rm B\kern-.05em{\sc i\kern-.025em b}\kern-.08em
    T\kern-.1667em\lower.7ex\hbox{E}\kern-.125emX}}
\usepackage{balance}

\usepackage[utf8]{inputenc} 
\usepackage[T1]{fontenc}    
\usepackage{caption} 
\usepackage[hidelinks]{hyperref}       
\usepackage{url}            
\usepackage{booktabs}       
\usepackage{nicefrac}       
\usepackage{microtype}      
\usepackage{xcolor}         

\usepackage{dsfont}
\usepackage{orcidlink}
\usepackage{color}
\usepackage{cleveref}

\begin{document}

\title{Grade Inflation in Generative Models

\thanks{
This work was supported by the Gordon and Betty Moore Foundation, the Food and Drug Administration, and the NIH under grants R01HL150394, R01HL150394-SI, 
R01AI148747, and R01AI148747-SI.}
\thanks{Phuc Nguyen, Miao Li, and Alex Morgan are with the Department of Pathology at Beth Israel Deaconess Medical Center (BIDMC), Boston, MA 02215. Rima Arnaout is with the Department of Medicine, the Bakar Computational Health Sciences Institute, and the UCSF UC Berkeley Joint Program for Computational Precision Health at the University of California San Francisco, San Francisco, CA 94143. Ramy Arnaout (email: rarnaout@bidmc.harvard.edu) is with the Department of Pathology and the Division of Clinical Informatics, Department of Medicine, BIDMC and with Harvard Medical School, Boston, MA 02215.}

}

\author{
    Phuc~Nguyen\orcidlink{0000-0001-9993-8434}, %
    Miao~Li\orcidlink{0009-0004-4951-2899}, %
    Alexandra Morgan\orcidlink{0000-0001-7787-0547}, %
    Rima~Arnaout\orcidlink{0000-0002-7134-0040}, %
    and~Ramy~Arnaout\orcidlink{0000-0001-6955-9310}%
}

\maketitle

\begin{abstract}
Generative models hold great potential, but only if one can trust the evaluation of the data they generate. We show that many commonly used quality scores for comparing two-dimensional distributions of synthetic vs. ground-truth data give better results than they should, a phenomenon we call the ``grade inflation problem.'' We show that the correlation score, Jaccard score, earth-mover's score, and Kullback-Leibler (relative-entropy) score all suffer grade inflation. We propose that any score that values all datapoints equally, as these do, will also exhibit grade inflation; we refer to such scores as ``equipoint'' scores. We introduce the concept of ``equidensity'' scores, and present the Eden score, to our knowledge the first example of such a score. We find that Eden avoids grade inflation and agrees better with human perception of goodness-of-fit than the equipoint scores above. We propose that any reasonable equidensity score will avoid grade inflation. We identify a connection between equidensity scores and Rényi entropy of negative order. We conclude that equidensity scores are likely to outperform equipoint scores for generative models, and for comparing low-dimensional distributions more generally.

\end{abstract}


\tableofcontents

\section{Introduction}

The ability to compare pairs of two-dimensional distributions robustly and accurately is critical in machine learning, and in data analysis more generally. This ability is especially valuable in the context of generative models \cite{miclaimgen} to measure model quality by assessing how well synthetic data fits training data. For example, generative modeling of tabular data, where each row is a datapoint and each column is a feature, has important uses across many fields \cite{lucie2024, beaulieujones2019, tkachenko2024, bauer_comprehensive_2024}. Generative modeling is of great interest for imaging and multi-modal data as well \cite{miclaimgen, arnaout2024, dey2023}. Although datasets for the latter applications are generally high-dimensional, two-dimensional analysis plays a critical role in quality assessment, either for testing the fidelity of pairwise relationships or via dimensionality reduction techniques such as PCA, tSNE, or UMAP  \cite{mcinnes_umap_2020, maaten_visualizing_2008, jolliffe_principal_2016}. 

Despite its ubiquity, the task of comparing two-dimensional distributions is non-trivial and has led to much work in developing quality scores. At a high level, quality scores can be classified as statistical vs. functional. Statistical scores are designed to demonstrate that some statistic, for example the mean of one of the features, has the same value in the synthetic dataset as in the real dataset (up to sampling error). In contrast, functional scores are meant to show that the outcome of some procedure, for example inference, is indistinguishable regardless of whether the input data is real or synthetic (again, up to sampling error) \cite{shmelkov_how_2018}. Statistical scores have the advantage of being easier to calculate and of being generalizable from dataset to dataset; functional scores often take more human and/or compute time and are more likely to be idiosyncratic to a particular dataset. For these reasons, statistical scores are of general interest.

Statistical scores appear often in the literature on generative modeling, especially of synthetic tabular data. For example, two widely-used and easy-to-compute statistics are the correlation coefficient $R$ and its square $R^2$ (the coefficient of determination), which measure the joint distribution between pairs of features (e.g. columns in tabular data). In turn, two common types of correlation coefficient are Pearson’s, which measures linear relationships, and Spearman’s, which is a generalization for any monotonic relationship. For a given pair of columns, one can calculate Pearson’s or Spearman’s $R$ for the real data ($R_p$), do the same for the synthetic data ($R_q$), and use them to calculate a \textit{correlation score}, for example as $1-|R_p-R_q|/2$ \cite{sdmetrics} or $1-|R^2_p - R^2_q|$. (Division normalizes the range to 0-to-1; subtraction converts the difference into a similarity score.) These two versions correlate closely with each other; the first is used more often in the literature, where it is also known as the correlation similarity.

\begin{figure}[t]
\centering
\includegraphics[width=0.5\textwidth]{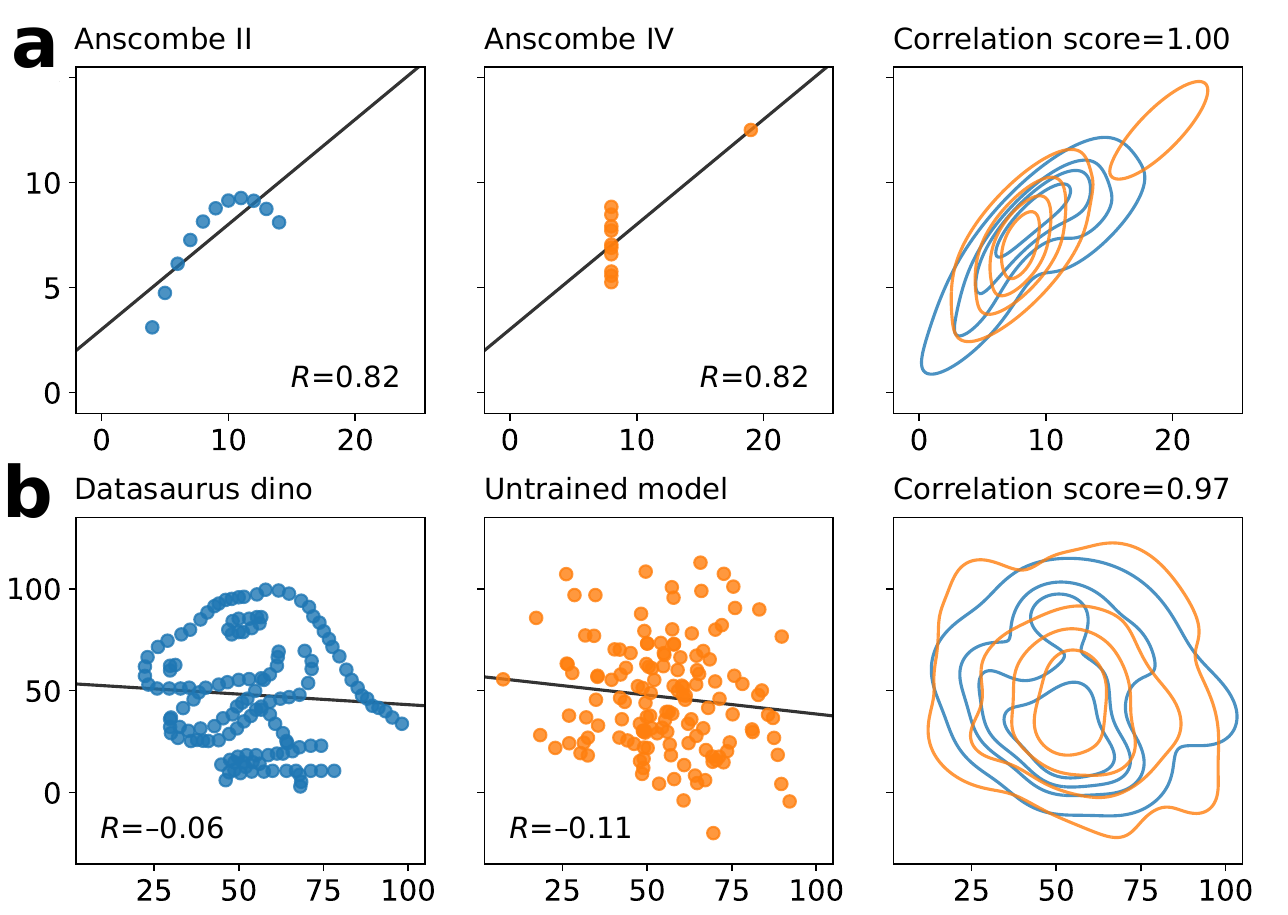}
\caption{The grade inflation problem. \textbf{a:} Two distributions from Anscombe's quartet \cite{anscombe_graphs_1973}. Both have a Pearson's $R$ of 0.82, meaning their correlation score is 1.00 despite their differences (which are appreciable in their KDEs, right). Black lines show least-squares regression fits, to illustrate indistinguishable slopes and intercepts. \textbf{b:} A highly non-random distribution from the Datasaurus Dozen \cite{matejka_same_2017} and (Gaussian-distributed) synthetic data with the same means and standard deviations from an untrained generative model. Pearson's $R$ of -0.06 and -0.11, respectively, resulting in the very high correlation score of 0.97 despite the poor fit.}
\label{fig:anscombe}
\end{figure}

There are two important challenges to using the correlation score as a measure of fit between a pair of two-dimensional distributions. The first is the well-known problem that many different distributions can have the same correlation coefficient, as illustrated by examples such as Anscombe's quartet \cite{anscombe_graphs_1973, matejka_same_2017} (Fig. \ref{fig:anscombe}a). Second, there are many more ways to get a low correlation than a high correlation. These two challenges can have the following effect. Suppose two columns in the real/training data have some non-random relationship that happens to have a near-zero correlation, for example like columns in the Datasaurus Dozen datasets, and that the synthetic/generated data fails to learn this relationship, resulting in a random distribution (Fig. \ref{fig:anscombe}b). Because random data has zero correlation (up to sampling variance), the correlation score between the real and synthetic data will be high, despite the low quality of the synthetic data. In principle such issues can to lead to inappropriately high quality scores, a ``grade inflation’’ problem that can make a generative model look better than it is.

To further characterize the grade inflation problem, here we evaluate several additional quality scores: the earth-mover's score, Jaccard score, and Kullback-Leibler or relative-entropy score. We find that all can fall prey to grade inflation. To address this problem, we introduce and describe a new score, the Eden score, that appears to avoid grade inflation. Our investigation focuses on two-dimensional distributions and pairwise relationships, which are often effective---sometimes ``unreasonably’’ so \cite{bialek_rediscovering_2007, schneidman_weak_2006}---at capturing key relationships in high-dimensional data.

\section{Methods}

\subsection{Datasets}
We generated synthetic data from both in-house two-dimensional toy datasets (called Dart, Trimodal, and Stripes) and the following high-dimensional machine-learning datasets obtained from the University of California Irvine machine-learning repository (UCIMLR): Covertype, Communities and Crime, Energy Efficiency, and Rice (Cammeo and Osmancik) (UCIML ID nos. 31, 183, 242, and 545, respectively). The latter were retrieved using the \textit{lucie} Python package \cite{lucie2024}. Anscombe's Quartet was obtained via the seaborn Python package. The Datasaurus Dozen were obtained via https://www.autodesk.com/content/dam/autodesk/www/autodesk-reasearch/Publications/pdf/SameStatsCode.zip from the file DatasaurusDozen.tsv.

\subsection{Generative models and KDEs}

The following models/model architectures were used to generate synthetic data: an in-house implementation of Gaussian Copula \cite{gaussiancopulareview}, the CTGAN implementation in the SDV package \cite{sdmetrics, xu_modeling_2019, xu_synthesizing_2020}, a flow-based model from the Python package nflow \cite{Stimper2023}, as well as in-house energy-based models \cite{couch_scaling_2023}. KDEs were generated using the kdeplot function of Python's seaborn package with levels=5 but otherwise default parameters \cite{waskom_seaborn_2021}. Note in seaborn the bandwidth of the KDEs is given by Scott's method \cite{scottsmethod}, and the lowest 5\% of the probability mass is ignored when finding the likelihoods of the contours.

\subsection{Correlation and earth-mover's scores}

We consider five different quality scores. The correlation score, the earth mover's score, the Jaccard score, and the Kullback-Leibler (KL) divergence score are based on previous work \cite{kullback_information_1951}; the Eden score is newly described in this work (see below). Correlation score is defined using Pearson's $R$ as $1-|R_p-R_q|/2$, with $p$ and $q$ the two distributions (e.g. real vs. synthetic data), reflecting the most common definition we found the literature. The earth-mover's score is defined as $e^{-k\cdot \mathrm{EMD}}$, where $k$ is a scaling factor that can be chosen to scale the score as desired (here, $k=1$), and EMD is the normalized earth-mover's distance from the pyemd python package \cite{pele2009}. Details for the remaining scores follow.

\subsection{Jaccard score}

The Jaccard score, also known as the intersection-over-union (IoU), is defined as
\begin{equation}\label{JaccardFormula}
J = \frac{|P \cap Q|}{|P \cup Q|}
\end{equation}
where the numerator is the size of the intersection between the two distributions $P$ and $Q$, and the denominator is the size of their union \cite{jaccard_distribution_1912}.

The size of the union is straightforward to define: it is the total number of points. In contrast, the intersection is a fuzzy notion due to points in the real and synthetic data generally being disjoint sets. Therefore, to define the intersection, we fit one Gaussian KDE model to the real data and a second Gaussian KDE to the synthetic data. We denote these real and synthetic KDEs $\hat{f}_{P}$ and $\hat{f}_{Q}$, respectively. We take the intersection to be:
\begin{eqnarray}
P \cap Q &=& \bigg \{ p \in P \bigg | \frac{\hat{f}_{Q}(p)}{\mathrm{max} \hat{f}_{Q}} > 0.1 \bigg \} \nonumber \\
&\cup& \bigg \{ q \in Q \bigg | \frac{\hat{f}_{P}(q)}{\mathrm{max} \hat{f}_{P}} > 0.1 \bigg \}
\end{eqnarray}
Thus, a point in the real data is in the intersection if its likelihood as evaluated by the synthetic KDE is sufficiently close to the maximum likelihood of the synthetic KDE (``sufficiently close'' is taken to mean that the ratio of the the two likelihoods $>0.1$), and similarly for the synthetic data.

\subsection{Kullback-Leibler score}

The KL divergence (relative Shannon entropy) is a special case of the Rényi divergence between two continuous distributions, defined as:

\begin{equation}\label{RényiDivergence}
D_{\alpha}{(p || q)} =\frac{1}{\alpha-1} \log{\left( \int p(\textbf{x})^{\alpha} q(\textbf{x})^{1-\alpha} d\textbf{x} \right)}
\end{equation}
Here, $\alpha$ is called the order parameter; the KL divergence is the Rényi diverence in the limit of $\alpha \rightarrow 1$. We defined the KL score as the exponential of the negative KL divergence:
\begin{equation}
\mathrm{KL~score} = e^{-D_{1}{(p || q)}}
\end{equation}
with $p$ and $q$ as above, approximated using Gaussian KDE also as above. Since the KL divergence takes values between $0$ and $\infty$, the KL score has the desirable range $[0,1]$. We approximated the integral for $D_{1}(p||q)$ by the Monte-Carlo method, which introduces (minor) stochasticity into the KL score. The Monte-Carlo approximation is as follows. We have:
\begin{equation}
D_{1}{(p || q)} = \int d\textbf{x} p{(\textbf{x})} \log{\left( \frac{p(\textbf{x})}{q(\textbf{x})} \right)}
\end{equation}
which can be rewritten as an expectation value:
\begin{equation}
D_{1}{(p || q)} = \mathbb{E}_{\textbf{x} \sim p} \left[ \log{\left( \frac{p(\textbf{x})}{q(\textbf{x})} \right)} \right]
\end{equation}
which in turn can be estimated from a sample of $p$:
\begin{equation}
D_{1}{(p || q)} \approx \frac{1}{N} \sum_{\textbf{x}_{i} \sim p} \log{\left(\frac{p(\textbf{x}_{i})}{q(\textbf{x}_{i})} \right)}
\end{equation}
where $N$ is the sample size.

\subsection{Eden score}

In the Eden score, the difference between two KDEs is calculated by calculating the difference for each successive ring or ``annulus'' defined by the contours of the KDEs, and averaging these differences. For each annulus in the real/synthetic data, we define a per-annulus similarity score. Without loss of generality, label the outermost annulus 0, the second-outermost annulus 1, and so on. (The unbounded region outside of all contours is excluded.) Define the $i^{th} $ per-annulus score by a variation of the Jaccard formula (Eq. \ref{JaccardFormula}), with the area playing the role of size:
\begin{equation}\label{Edeni}
s_{i} = \frac{\mathrm{Area}(\mathrm{i^{th}~annulus~of~}p \cap \mathrm{i^{th}~ annulus~of~}q)} {\mathrm{Area}(\mathrm{i^{th}~annulus~of~}p \cup \mathrm{i^{th}~annulus~of~}q)} 
\end{equation}
The Eden score is obtained by averaging scores over all annuli:
\begin{equation}
\textrm{Eden} = \frac{1}{n_{\text{annuli}}} \sum_{i=0}^{n_{\text{annuli}}-1} s_{i}
\end{equation}
Here $n_{\text{annuli}}$ is the number of annuli. Note, technically the innermost annulus is usually a disk, and any annulus can consist of non-contiguous densities (e.g. multiple peaks). $n_{\text{annuli}} = 5$ was used.

To estimate the areas in the numerator and denominator of Eq. \ref{Edeni}, a Monte-Carlo method is used: a bounding rectangle centered at the data is sprinkled with a large number of uniformly distributed points; the number of points that lie inside the union/intersection of a real and a synthetic annulus $i$ is counted. To determine whether a point lies inside a union/intersection of two annuli, we compute the likelihood of the point under $\hat{f}_{r}$ and $\hat{f}_{s}$ and check whether the likelihood falls within the likelihood range that defines two adjacent contours. As with the KL score, this process introduces minor stochasticity. 

\subsection{Validation}

We used human visual inspection as a gold standard for comparison of two-dimensional distributions. 39 pairs of plots were shown to 20 human raters. Raters all had a background in data science, science, and/or engineering, to increase the likelihood of exposure to/familiarity with the general practice of data presented as KDEs. Each plot consisted of two superimposed KDEs in different colors, corresponding to training data and synthetic data output by a generative model. Plots were chosen so that the plot with the higher Eden score received a lower score according to at least one of the other scores. Scores were not shared with the human raters, making this a blind test. 

Each human rater was asked to choose the plot in which the contours matched better, considering all contours; the interpretation of ``better'' was otherwise left up to the rater. The 39 pairs corresponded to 3 repeats of each of 13 unique pairs of columns. For each repeat, the pair was subjected to rotations or color swaps, and the left-right order of the plots was randomized; this allowed for a per-person test of consistency, to assess whether the rater picked the same plot all three times regardless of position, orientation, and color. Separately (i.e. without the rater), the fit in each plot was scored according to each scoring method---correlation, earth-mover's, Jaccard, KL, and Eden---to determine the higher-scoring plot in each pair. Cohen's kappa was used to calculate the agreement between each rater and each scoring method. The Mann-Whitney U test was used to test the null hypotheses that each two scoring methods were equally good (Python, scipy.stats.mannwhitneyu).

\section{Results}

\subsection{Correlation score}
To illustrate how the correlation score can lead to grade inflation, we first measured this score between pairs of distributions from Anscombe's quartet. The four distributions in this set were devised in the 1973 to show how different distributions can have identical Pearson's $R$ (to several digits). Fig. \ref{fig:anscombe}a illustrates the outcome: a perfect correlation score, despite the two distributions differing materially from each other. This is grade inflation. The ``dino'' dataset from the Datasaurus Dozen illustrates the compounding problem that arises as a result of low-correlation distributions being more common than high-correlation ones (Fig. \ref{fig:anscombe}b). Specifically, treating dino as the training set, we randomly intialized a generative model with the same x and y means and standard deviations, and sampled from that model without any training. The dino dataset has a Pearson's $R$ of close to zero. Random data also has a Pearson's $R$ of close to zero. Because these values are similar, the correlation score is nearly perfect---0.97 in the sample in Fig. \ref{fig:anscombe}b---despite the model having learned literally nothing beyond the location and scale of the data. Again, this is grade inflation.

\begin{figure}[t]

\includegraphics[width=0.5\textwidth]{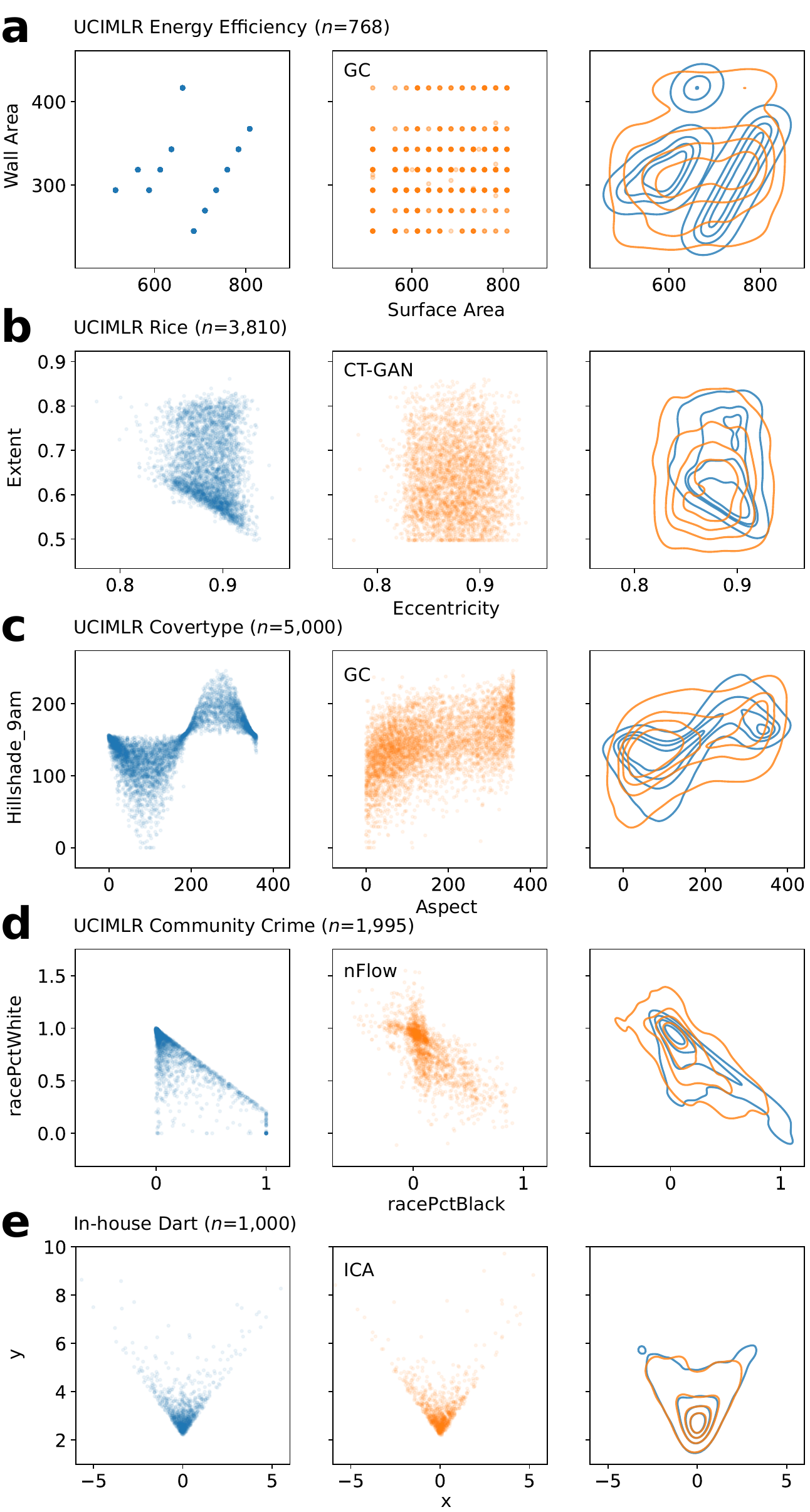}

    \caption{Fits scored in Table \ref{table:scores}. Left to right: real data (blue), generated synthetic data (yellow), and KDEs. \textbf{a} is considered a very low-quality fit; \textbf{b-d} are considered low-quality fits; \textbf{e} is considered a high-quality fit. Datasets, sizes, features, and models are as labeled. UCIMLR=UC Irvine Machine-Learning Repository. GC=Gaussian copula. nFlow=normalizing flow. ICA=independent component analysis.}
\label{fig:fits}
\end{figure}

To illustrate the phenomenon on datasets and generative models used in real-world data science, we applied a variety of models to a selection of datasets from the UCIMLR and measured the correlation score between the synthetic/generated data and the real/training data, for representative pairs of columns (Fig. \ref{fig:fits}a-d). Even when fits were low quality by eye, the correlation score was universally high, ranging from 0.903 to 0.994 (Table \ref{table:scores}), demonstrating grade inflation. As a positive control, we also scored a high-quality model fit of an in-house dataset called Dart (Fig. \ref{fig:fits}e). Not surprisingly, the correlation score was also excellent, at 0.981, but notably this score was actually \textit{lower} than the 0.994 achieved by the low-quality fit in Fig. \ref{fig:fits}a (Table \ref{table:scores}). Thus, correlation score has difficulty differentiating between high- and low-quality fits, leading to grade inflation for some of the latter.

\subsection{Earth-mover’s score}
A more sophisticated score derives from the earth-mover's distance (EMD; a.k.a. the Mallow, Wasserstein, or Kantorovich-Rubenstein distance) \cite{vaserstein1969markov, kantorovich_mathematical_1960, rubner_metric_1998, mallows_note_1972}, which is the basis of the Frechet inception distance that is commonly used in machine learning \cite{heusel2018}. If each two-dimensional distribution is a pile of sand, the EMD is the minimum amount of work required to transform one distribution into the other by moving sand around. EMD is easily converted to a score with range $(0, 1)$ via exponentiation (Methods). To apply this score, the finite collections of datapoints in each distribution are binned into histograms. (Note, the algorithm used here, from \cite{pele2009}, does not require smoothing the data by KDE, though that is an alternative approach.) In its native form, the EMD has the undesirable feature of being sensitive to the overall number of data points; to avoid this scale-dependence, the real and synthetic data are generally normalized, as we do here, before computing the score.

We calculated the earth-mover's score for each fit in Fig. \ref{fig:fits} and found this score also exhibits grade inflation (Table \ref{table:scores}). Interestingly, it resulted in high scores in all cases where the training data exhibited sharp boundaries, regardless of whether the fit was of low quality, as in Figs. \ref{fig:fits}b and d, with scores of 0.982 and 0.959, respectively, or high quality, as in Fig. \ref{fig:fits}e, which it gave a score of 0.949. The earth-mover's score did, however, result in (appropriately) low scores in the low-quality fits of the two multimodal distributions, Figs. \ref{fig:fits}a and c, with scores of 0.124 and 0.379, respectively. Of note, scaling the scores by choosing a different value of $k$ does not resolve the grade inflation in Figs. \ref{fig:fits}b and d, because scaling so that lower quality fits have lower scores has the adverse effect of also reducing the score of high-quality fits: because changing $k$ does not change the ordering of scores, $k$ cannot thread this needle. Also of note, the earth-mover’s score can be highly sensitive to outliers, because the optimal transport might be such that the outliers have to be moved over further distances compared to more central points (see Discussion).

\begin{table*}[tb]
\begin{center}

\includegraphics[width=0.95\textwidth]{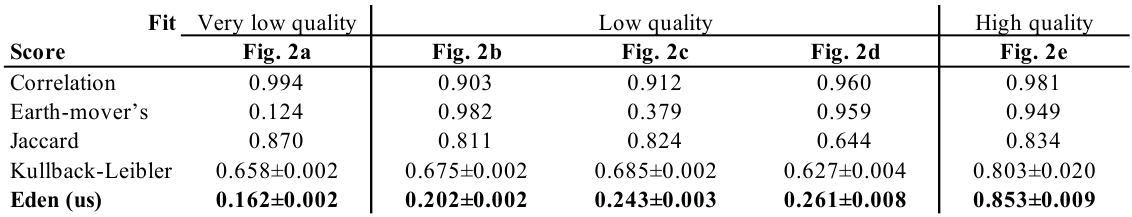}
\end{center}
\caption{Scores for the fits and samples in Fig. \ref{fig:fits}. Fig. \ref{fig:fits}a-d are considered very low- or low-quality fits deserving lower scores; Fig. \ref{fig:fits}e is considered a high-quality fit deserving higher scores. See Results for details. Standard deviations are reported for scores whose calculation involves a stochastic element (Methods); they are not for repeat sampling (see Confidence Intervals).}
\label{table:scores}
\end{table*}

\subsection{Jaccard score}
A score often used in the evaluation of generative models is the Jaccard score or intersection-over-union (IoU) \cite{jaccard_etude_1901, jaccard_distribution_1912}. It has the desirable range of 0 to 1. Of note, the Jaccard score does not exhibit the outsize sensitivity to outliers that the earth-mover's score has, since the relative distance between non-intersecting densities does not enter into the equation (Methods). In our analysis, the Jaccard score was found to behave like the correlation score in that values were similar regardless of whether fits were high- or low-quality (Table \ref{table:scores}). The main differences were (1) the range of scores was lower than for the correlation score and (2) one of the low-quality fits---Fig. \ref{fig:fits}d, with a Jaccard score of 0.644---scored ~0.2 lower than the others, which ranged from 0.811 to 0.870. Importantly, the score of 0.834 for the high-quality fit (Fig. \ref{fig:fits}e) was actually lower than for the lowest-quality fit (0.870 for Fig. \ref{fig:fits}a), indicating grade inflation. We found that the Jaccard score is especially prone to grade inflation when one distribution completely circumscribes the other. Together, these results support the conclusion that, like the correlation and earth-mover's scores, the Jaccard score is not a dependable discriminant of quality in generative models, due to grade inflation.

\subsection{Kullback-Leibler score}
A third relevant quality score is based on the Kullback-Leibler (KL) divergence, also known as relative entropy. This quantity is interpreted as the information lost, or the ``surprise,'' in approximating one distribution by another \cite{leinster_entropy_2024, callum2022}. It is used widely in generative models, although in our reading of the literature not as frequently for evaluating synthetic data as the correlation, earth-mover's, or Jaccard scores. However, the ubiquity of the KL divergence and its utility comparing two-dimensional distributions make it a natural comparator.

KL divergence is one of a family of Rényi divergences parameterized by $\alpha$, the order. From Eq. \ref{RényiDivergence}, $\alpha$ weights the extent to which the density of $p$ contributes to $D_{\alpha}$. For $\alpha \rightarrow \infty$, $D_\alpha$ is almost completely determined by the highest-density region of $p$. In contrast, for $\alpha \rightarrow -\infty$ (the most extreme example of negative order), the sign flip results in $D_\alpha$ being almost completely determined by the highest-density region of $q$; i.e. the roles of $p$ and $q$ are essentially swapped ($D_\alpha$'s sign is also flipped). Meanwhile, $\alpha = 0$ means density is not weighted at all, resulting in the trivial result of $D_{\alpha} = 0$ for all $p$ and $q$ (assuming $p$ and $q$ do not vanish anywhere): this is because any distributions $p$ and $q$ are trivially indistinguishable if one ignores their density distributions. KL divergence corresponds to $\mathrm{lim}(\alpha \rightarrow 1)$. The divergence is easily converted into a score by exponentiation (Methods).

While the KL scores from Table \ref{table:scores} are not that high in absolute terms, the fit for Fig. \ref{fig:fits}a suffers grade inflation relative to the fits for Figs. \ref{fig:fits}c-d, which are better fits by eye but have similar scores. The reason for the grade inflation in this case is because the real and synthetic data have similar support but different numbers of modes \cite{callum2022}, as follows. Recall that KL divergence can be understood as the amount of surprise at the real data, given the synthetic data. In Fig. \ref{fig:fits}a, the synthetic data has a single mode, with a support that spans the the real data's two main modes. Even though the fit is low quality, it does have density that overlaps the centers of the two main modes in the real data. As a result, the surprise factor at finding density in these regions of the real data is low, inflating the KL score (which varies inversely with the divergence), despite the low-quality fit. This inconsistency is a drawback of the KL score.

\subsection{Eden score} 
\begin{figure}[t]
    \centering
    \begin{center}
        \includegraphics[width=0.49\textwidth]{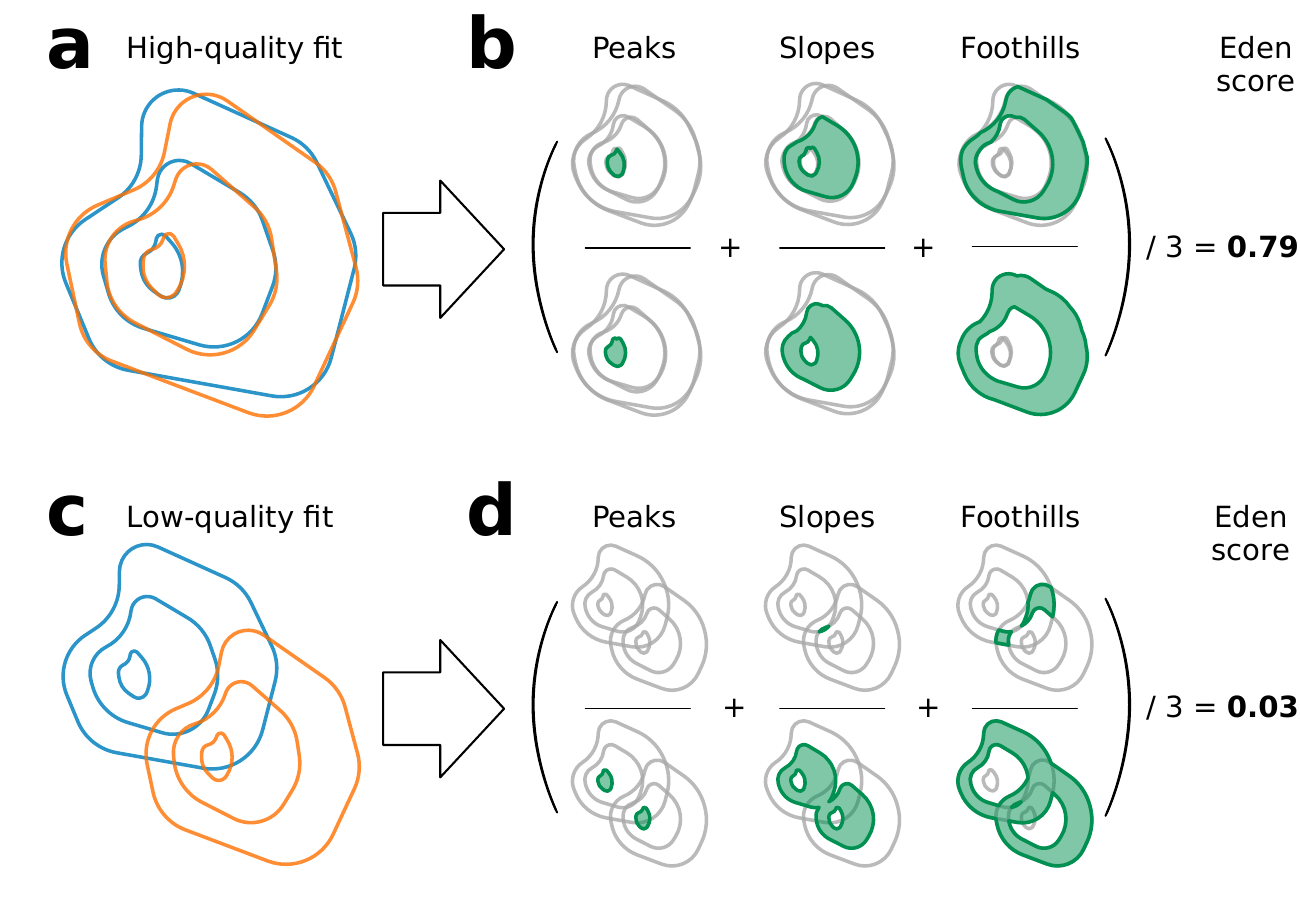}
        \caption{The Eden score. The Eden score comparing the blue and yellow distributions in the high-quality fit in (\textbf{a}) is calculated as the mean intersection-over-union for each equidensity contour (ring, annulus) (\textbf{b}). Peaks, slopes, and foothills contribute equally. Top row; intersections; bottom row, unions (both in green). The ratios for the three contour levels shown are 0.77, 0.81, and 0.78 (left-to-right), which average to an Eden score of 0.79 for the fit of the two distributions in (a). For clarity, the score is calculated over three contours, instead of the five used in the rest of this study. \textbf{c-d:} Similar for a low-quality fit. The peaks are disjoint (ratio, 0.00), the slopes intersect by only a sliver (ratio, <0.01), and the foothills' intersection-over-union is 0.09, yielding an Eden score of 0.03.}
    \label{fig:eden}
    \end{center}
\end{figure}

The Eden score differs qualitatively from the correlation, earth-mover's, Jaccard, and KL scores in being an \underline{e}qui\underline{den}sity score, whence its name (see Discussion). Eden is based on the principle that a distribution $p$ is a good match of a distribution $q$ if their regions of probability density $i$ coincide, for all densities $i$. Any reasonable method can be used for determining coincidence; we use the Jaccard score, since Jaccard exhibits grade inflation only when densities vary, which is definitionally not the case for equidensity regions; therefore unlike when comparing complete distributions, comparison of equidensity regions should not have this problem. To calculate the Eden score, the Jaccard score is calculated for each annulus or ring of density $i$ and these scores are then averaged (Fig. \ref{fig:eden}). All contours of the distribution count equally, avoiding the issues encountered with the earth-mover’s and Jaccard scores (and, of course, with the correlation score). 

Indeed, calculation of the Eden score for each of the fits in Fig. \ref{fig:fits} showed that it avoids grade inflation (Table \ref{table:scores}). Scores for the low--quality fits in Fig. \ref{fig:fits}a-d ranged from 0.162 to 0.261 vs. 0.853 for the high-quality fit in Fig. \ref{fig:fits}e. (The confidence intervals in Table \ref{table:scores} account for the stochastic element in our implementation, which can be made arbitrarily small by scaling up the Monte Carlo; the same is true for the KL score. The others are deterministic.) Eden was the only score to demonstrate a consistent gap between low- and high-quality fits, without exceptions. Moreover, this gap was sizable, at 3-4x. Thus, it was immune to grade inflation in these examples.

\subsection{Validation}
Human visual inspection leverages the visual cortex's millions of years of evolved expertise at comparing sizes and shapes to provide a gold-standard assessment of the similarity of two-dimensional distributions. Quantitative expertise hones this ability. Therefore, to validate our results, we showed 20 scientists, data scientists, and engineers 13 fits from pairs of generative models and asked them to choose the better fit. We then compared their choices to each of the five scores we evaluated: correlation, earth-mover's, Jaccard, KL, and Eden. Raters were not shown any scores, making this a blind test.

\begin{figure}[t]
    \centering
    \begin{center}
    \includegraphics[width = 0.48\textwidth]{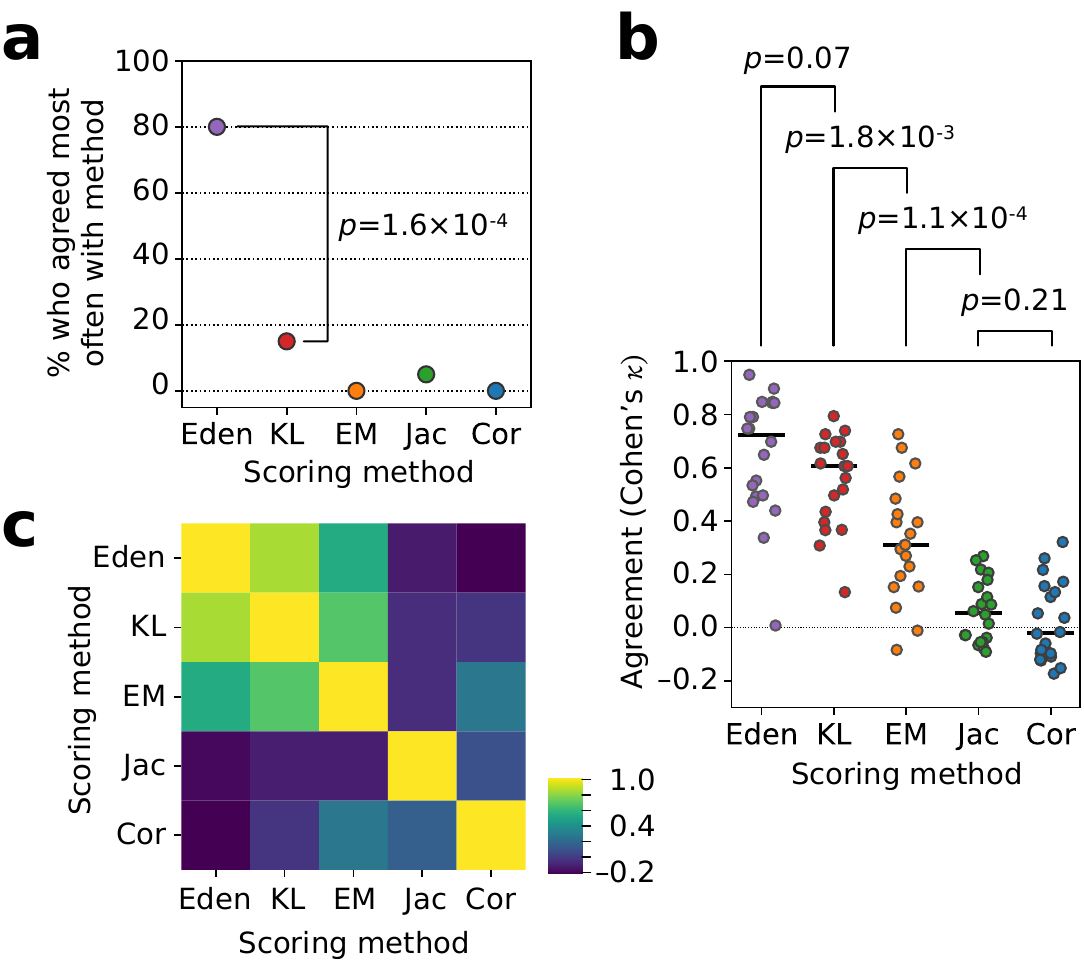}
    \caption{Validation. 20 human experts were each asked to rate which of two fits was better for several pairs. Ratings were then compared against each of the five statistical scoring methods. Agreement between each human rater and the scoring method was measured by Cohen's $\kappa$. \textbf{a:} Percent of raters who agreed the most with each scoring method (Eden, KL, etc.). p-value is for Mann-Whitney U on the ranks. \textbf{b:} All $\kappa$ values for each score, with Mann-Whitney U p-values (Methods). \textbf{c:} Agreement between methods for test pairs (again measured by $\kappa$).}
    \label{fig:blind}
    \end{center}
\end{figure}

 Eighty percent of raters agreed most closely with the Eden score (Fig. \ref{fig:blind}a), vs. 15\% for KL (MWU $p=1.6\times10^{-4}$), 5\% for Jaccard, and 0\% for the earth-mover's and correlation scores. Although Eden consistently outperformed KL on a person-by-person basis (Fig. \ref{fig:blind}a), its advantage over KL was small overall (Fig. \ref{fig:blind}b), with a median Cohen's $\kappa$ of 0.722 across raters (the ``excellent'' range; 10th--90th percentile: 0.429--0.852) vs. 0.606 (``moderate'' agreement; 0.360--0.727) for KL. The explanation: Eden and KL happened to agree on most pairs of fits in the test (Fig. \ref{fig:blind}c), an artifact of how the pairs were chosen (Methods). When the analysis in Fig. \ref{fig:blind}b was repeated on only those pairs on which Eden and KL disagreed (not shown), 80\% of raters agreed with Eden over KL, consistent with Fig. \ref{fig:blind}a. 
 
 Both Eden and KL substantially outperformed the earth mover's score (median $\kappa=0.311$; ``low'' agreement, 10th-90th percentile 0.129--0.634), correlation score (median $\kappa=-0.019$; -0.126--0.221), and Jaccard score (median $\kappa=0.055$; -0.067--0.222). Of note, the near-zero median $\kappa$s of the correlation scores and Jaccard scores indicate that agreement between these scoring methods and humans with domain expertise is no better than chance, arguing against their use. In contrast, Eden's showed excellent agreement with human gold standard. Together with its avoidance of grade inflation, including both its clear segregation between low- and high-quality fits and its scoring consistency, Eden's performance supports its use for evaluating two-dimensional distributions.

\subsection{Confidence intervals}
\begin{figure}[t]
    \centering
    \begin{center}
    \includegraphics[width = 0.50\textwidth]{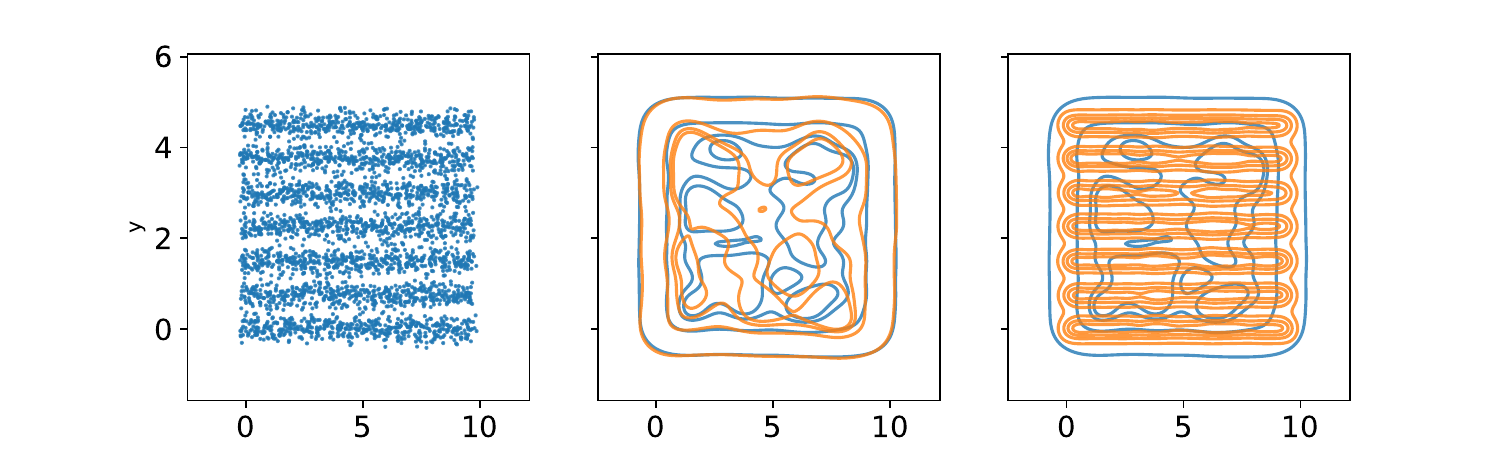}
    \caption{Oversampling affects scoring. \textbf{a:} Target (in-house ``Stripes'' dataset). \textbf{b:} Sample (orange) the same size as the target. Correlation, earth-mover's, Jaccard, KL, and Eden scores: 0.993, 0.941, 1.000, 0.996, and 0.452, respectively. \textbf{(c)} Oversampling. Scores (same order): 0.998, 0.992, 1.000, 0.862, and 0.160. Eden is the most sensitive to the difference in KDEs between normal sampling (b) and oversampling (c).}
    \label{fig:stripes}
    \end{center}
\end{figure}

 For the most direct, apples-to-apples comparisons, generated synthetic data should have the same number of datapoints as the training data. However, synthetic data is often needed precisely because the training dataset is small; in principle, sampling the equivalent small number of synthetic datapoints can result in comparatively large sampling error in quality scores. However, we found that the alternative of oversampling from the model can result in a synthetic-data KDE that is qualitatively sharper than that of the real dataset, potentially complicating comparisons (Fig. \ref{fig:stripes}). To illustrate, we trained a Gaussian copula on a target with a striped pattern (Fig. \ref{fig:stripes}a). The normal KDE plot (i.e. without oversampling of the synthetic data) does not pick up the stripes (Fig. \ref{fig:stripes}b), whereas the KDE plot with oversampled synthetic data does (Fig. \ref{fig:stripes}c). This lowers the Eden score (from 0.452 to 0.160), as might be expected from how different the KDE becomes. Of the five scoring methods, only Eden and KL were sensitive enough to fall in response to this difference. The correlation, earth-mover's, and Jaccard scores actually rose, but were all $\geq$0.941 to begin with, reflecting the grade inflation problem.

\begin{figure}[t]
    \centering
    \begin{center}
    \includegraphics[width = 0.48\textwidth]{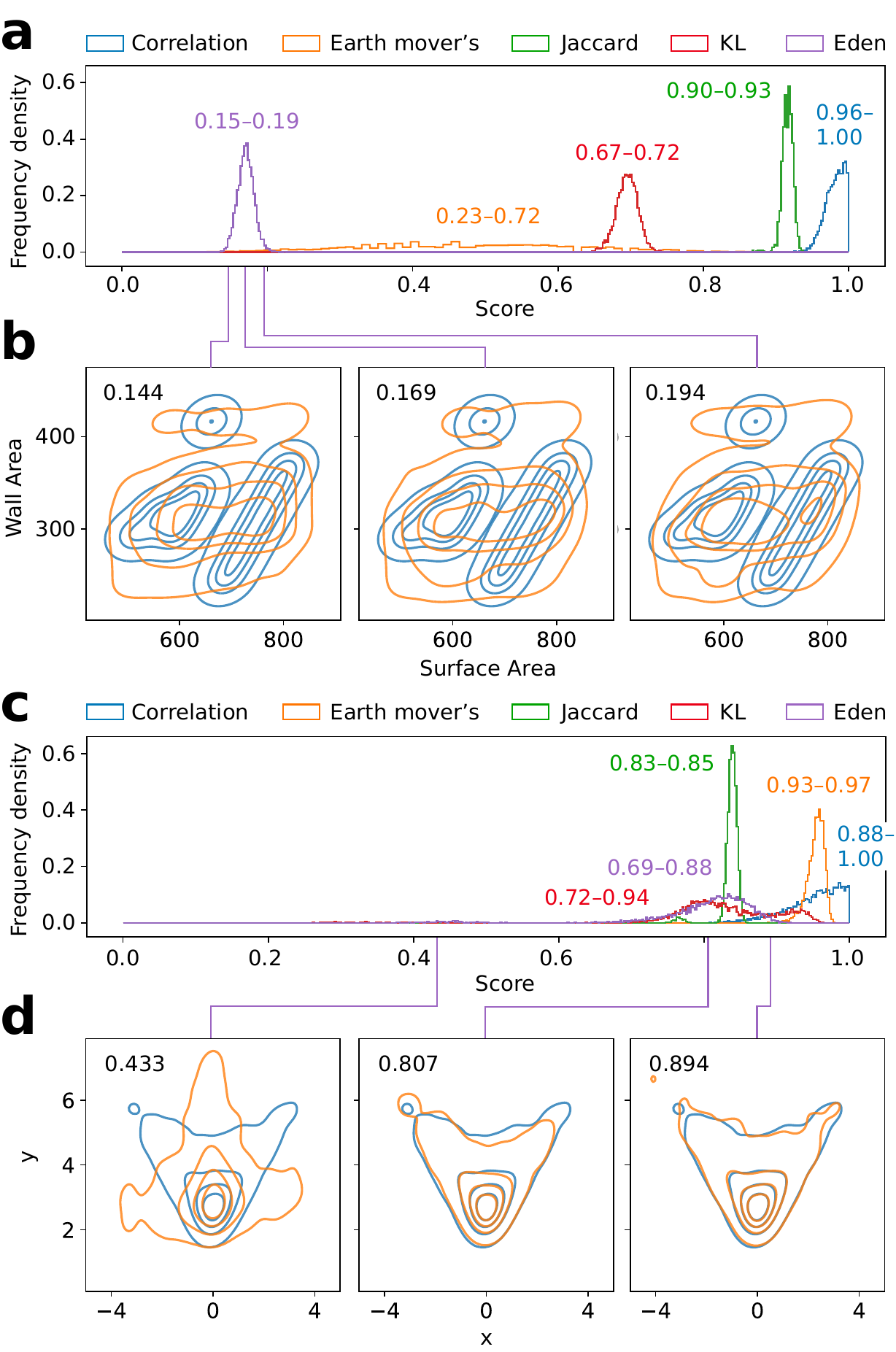}
    \caption{Confidence intervals. \textbf{a:} Score distributions from 5,000 repeat samples for the fit from Fig. \ref{fig:fits}a, a very low-quality fit; 5th- and 95th-percentile limits shown. Note, all 5,000 samples are from the same fit. \textbf{b:} Representative KDE plots for samples drawn (purple lines) from different parts of the Eden score distribution (with Eden scores shown). \textbf{c:} Score distributions from 5,000 repeat samples for the fit from Fig. \ref{fig:fits}e, a high-quality fit, with 5th- and 95th-percentile limits. Again, all 5,000 samples are from the same fit. \textbf{d:} Representative KDE plots for samples drawn from different parts of the Eden score distribution (with Eden scores shown).}
    \label{fig:CIs}
    \end{center}
\end{figure}

Our results support repeat sampling as a middle ground between sampling error and mismatch due to oversampling: i.e., scoring each sample and reporting summary statistics such as the mean ± standard deviation of the scores, the median, interquartile scores, or simply the $n$th percentile score. The latter is more conservative than the mean or median: it is the score that $n$ percent of the scores are better than. Figs. \ref{fig:CIs}a and \ref{fig:CIs}c show the distribution of values for each score for 5,000 resamplings for the very low-quality fit from Fig. \ref{fig:fits}a and the high-quality fit from Fig. \ref{fig:fits}e, respectively. (Figs. \ref{fig:CIs}b and \ref{fig:CIs}d show KDEs from different samples drawn from the range of scores for Eden.) Notably, the score distributions spanned at least several percent for all five scoring methods. The earth-mover's score on the very low-quality fit was especially variable, with an interquartile range of $0.22$ (5th-95th percentiles, $0.23$-$0.72$). The extent of the variation observed in Fig. \ref{fig:CIs} supports caution in grading generative models, and strongly suggests that results from single samples can be misleading, may lack discriminative power, and therefore should be avoided if possible.

\section{Discussion}

All modeling benefits from reliable quality scores, including two-dimensional distributions resulting from generative models. Scores can mislead in several ways, including overfitting and in leading to selection of the wrong model \cite{greene_model_2022}. Here we add to these dangers the grade-inflation problem, named for a perenially-decried phenomenon in U.S. higher education \cite{Wikipedia_gradeinflation}, in which a statistic scores a model higher than it should. We describe why the commonly used correlation score should be particularly prone to grade inflation, and offer examples from real-world datasets where other commonly used scores---specifically, the earth-mover’s, Jaccard, and KL scores---still have this problem, while at least one other score, our newly proposed Eden score, appears not to. We also show the value of multiple sampling for measuring and reporting confidence in these scores, which appears to be a relatively uncommon practice in the literature. We suspect the grade-inflation problem is not new, but is made newly relevant by the explosive growth of data science, the need to select among high-performing models, the inability to keep up via human visual inspection, and the computational resources increasingly available for data visualization (for example, enabling pairwise scatter plots for ever higher-dimensional datasets).

It is interesting that the Eden score agrees substantially better with human perception of goodness-of-fit than the other scores tested, in a blind head-to-head comparison on data from generative models. This finding suggests that when human raters compare distributions for similarity, they, like the Eden score, might also be comparing distributions at multiple densities and subconsciously averaging the result. The caveats: raters were specifically asked to consider all contours, were shown the distributions as KDE plots (as opposed to, for example, as three-dimensional and/or interactive renderings), and were limited to people with scientific backgrounds. This is an interesting topic for further investigation.

\subsection{Equipoint vs. equidensity scores}
What explains the difference between the earth-mover's, Jaccard, and KL scores on the one hand, which exhibited grade inflation, and Eden, which did not? One answer is that these two sets of scores differ qualitatively in how they weight different regions of the two distributions being compared. The earth-mover’s score weights each datapoint equally, regardless of where in a distribution that point lies. Because by definition there are more datapoints in areas of higher density, the earth-mover’s score will tend to be high as long as the regions of highest density line up well between the two distributions, almost regardless of how mismatched the low-density regions are (absent extreme outliers). The Jaccard and KL scores also weight each datapoint equally, with similar results. Note, these three scores are likely not the only ones with this property. For example, multi-dimensional extensions of the Kolmogorov-Smirnov [KS] statistic also seem to have this issue \cite{harrison_validation_2015}, and the Frechet inception distance is likely to suffer grade inflation as well, since it is based on the earth-mover's distance. We see no obvious reason that scoring methods based on other $f$-divergences would not also have this problem. We coin the term ``equipoint'' for scores of this kind. Based on these examples, we hypothesize that all equipoint scores will suffer from this tyranny of the majority, yielding high scores even when low-density regions are poorly fit, resulting in grade inflation (Fig. \ref{fig:fits} and Table \ref{table:scores}).

In contrast, the Eden score weights \textit{datapoints} unequally, such that all \textit{densities} contribute equally to the score. In such ``equidensity'' scores, the region of lowest density carries equal weight to the region of highest density, and indeed to all density regions in between (Fig. \ref{fig:eden}). Topologically, both the peaks and the foothills of the real and synthetic distributions have to line up well to get a high equidensity score, whereas only the peaks need to line up to get high equipoint scores. We propose that equidensity scores such as Eden should be preferred whenever accuracy is required throughout the distribution. (We exclude ``unreasonable'' equidensity scores from consideration, e.g. taking a trivially large $n$th root of Eden such that all scores end up arbitrarily close to 1.) For example, for medical applications, it is often just as important that generative models accurately capture the features of rare presentations (foothills) as common ones (peaks) \cite{dey2023}. Since one of the main purposes of generative models is to fill out rare cases, we conclude that equidensity scores will generally have the advantage over equipoint scores for this application. Although to our knowledge the Eden score is the first equidensity score, or at least the first recognized as such, it is possible that scores based on other measures that up-weight tails also have something of an equidensity character; potential examples include the one-dimensional Anderson-Darling statistic, which up-weights tails more than the KS and e.g. Cramér-von Mises statistics, and which can be extended to two dimensions \cite{anderson1952}. This is a topic for future study.

\subsection{Score calibration}
Like all scores, equidensity scores can be calibrated to a desired range of values, for example by raising to an exponent (for scores that range from 0 and 1). Importantly, note that for a given scoring statistic, calibration cannot change which of two fits has the higher value. For example, no calibration procedure can make the correlation score for Fig. \ref{fig:fits}e larger than for Fig. \ref{fig:fits}a (Table \ref{table:scores}). This is a structural problem with the correlation score; as we have shown, it affects the equipoint scores as well. This means that even though all correlation scores can be brought arbitrarily low in hopes of resolving the \textit{absolute} grade inflation they exhibit, for example by raising the correlation score to a sufficiently large exponent, structurally the correlation score will still always exhibit \textit{relative} grade inflation, in which low-quality fits will perform as well as or better than high-quality ones. Thus, calibration cannot fix grade inflation. The goal of the present work has been to investigate scores' structural properties. Calibration of the Eden score or other scores, if desired, is left for future work.

\subsection{Sampling error and confidence}
It is interesting to find that confidence intervals were not negligible (Fig. \ref{fig:CIs}). In several of the examples we investigated, even the inter-quartile ranges, much less the 5th-95th percentile ranges, exceeded the differences between mean values for scores on the example fits in Fig. \ref{fig:fits}. This observation is potentially important for comparing models: if the score used is one whose confidence interval is comparable to or even wider than the difference in score values between two models, one might mistake the worse model for the better one, or conclude a difference when no statistically significant difference exists. We note that in the machine-learning literature, model performance is often compared out to several decimal places without confidence intervals being reported. Fig. \ref{fig:CIs} suggests interpreting such results with caution.

Consideration of variance raises the question of whether the Eden score should always be calculated with five annuli. We propose the choice should depend on the total number of datapoints, such that sufficiently many points lie in each annulus to provide for acceptable sampling variation and acceptable confidence intervals. Because sampling error scales as the root of the number of points, we suggest one approach is to choose the number of annuli so that there are at least roughly 30 points per annulus; more points per annulus will result in narrower confidence intervals, but more annuli will result in a higher-resolution comparison of the two distributions. We believe this is essentially a precision-accuracy trade-off that is the choice of the investigator. (In this study we used five annuli throughout, to control for any effects variation in this choice might have.) For general applications beyond synthetic data in which the number of points is non-limiting---for example, when the two distributions are continuous and defined at every point---it could be interesting to define a continuum limit for $n_{\text{annuli}} \rightarrow \infty$. This is left for future work.

\subsection{Connections to entropy and diversity}
The correspondence between unequal weighting at the level of datapoints and equal weighting at the level of densities has an interesting connection to entropy and diversity, specifically to the Rényi entropies $H^\alpha$ and the corresponding Hill diversities $D^q=\exp H^\alpha$. (Here $\alpha=q$, a potentially confusing but purely notational difference reflecting the conventions in the respective literatures; here we distinguish $q$ the viewpoint parameter from $q$ the distribution by context.) Both $H^\alpha$ and $D^q$ can be interpreted as sums of the frequencies of \textit{species} in a system, with $q$ as a frequency-weighting parameter such that rarer species contribute less as $q$ rises. $D^q$ is used to calculate the \textit{effective number} of species in a population, taking frequencies into account to a degree $q$. It has been observed \cite{hill_diversity_1973, jost_entropy_2006, leinster_entropy_2024} that many commonly used statistics correspond to positive integer values of $q$. For example, $q=0$ corresponds to a simple count of the number of unique species, $q=1$ corresponds to the Shannon entropy, $q=2$ corresponds to Simpson’s index, and so on up to $q=\infty$, which corresponds to the Berger-Parker index \cite{berger_diversity_1970}. (These correspondences generally take the form of simple mathematical transformations of $D^q$.) 

$q$ ($=\alpha$) can also take negative values; however, entropies/diversities with negative $\alpha$ have received little if any attention in the literature, perhaps owing to a dearth of real-world examples for $\alpha<0$ \cite{li_renyi_2016}. We broach the possibility that equidensity scores might be interpreted as an example of negative $\alpha$ if one considers the species in the entropy calculation to be \textit{datapoints}, and that there is a duality with $\alpha=0$ if instead one considers the species to be \textit{equidensity regions} (i.e., annuli/rings/topological contours). Note this is qualitatively different from how negative $\alpha$ operates in the Rényi divergences, which is to swap distributions $p$ and $q$ (to a reasonable approximation, increasingly true the further from $\alpha=0$). Insofar as equidensity regions are groupings or ``communities’’ of datapoints, there may also be connections with the subcommunity/metacommunity formulation in the diversity literature \cite{reeve_how_2016}, which correspond to concepts such as relative/joint entropy and mutual information in the entropy literature \cite{leinster_entropy_2024}. We note that a Python package already exists that accepts negative $\alpha$ \cite{greylock}. Exploring such connections could be an interesting direction for future work.

\subsection{Use cases and exceptions}
The general use case for the Eden score is to compare pairs of two-dimensional, i.e. pairwise, distributions. As we have shown, this makes it highly applicable for generative modeling, and for two-dimensional pairwise comparisons more generally. We propose that Eden could also be useful for comparing pairs of three-dimensional distributions, but caution that comparisons in still-higher dimensions could incur the curse of dimensionality.

The question arises of whether there two-dimensional cases where Eden might not be preferred. We believe there are at least two such cases. The first is where there is a strong prior on the two-dimensional data, for example when the expectation is that the data follows a simple line or curve. For example, suppose each of the two distributions is expected to lie on a simple straight line. In such cases, a regression-based score such as the correlation score can potentially still be useful; however, as Anscombe showed, it is still worth approaching with caution \cite{anscombe_graphs_1973}.

The second case where Eden might not be preferred is when the data configuration is very highly constrained. For example, in the extreme case where each feature is a binary variable (Boolean data), the space of possibilities consists of only four possible points: the coordinates (0,0), (0,1), (1,0), and (1,1). This constraint leaves no meaningful density regions as such. As a result, equidensity scores are unlikely to be of benefit. In such cases, an alternative approach is to treat the four coordinates as categoricals, and score agreement between the resulting two distributions accordingly. In contrast, Eden is for the general use case of complex and/or nonparameteric data defining meaningful densities in two dimensions. Potential additional use cases and limitations are left for future investigation.

\subsection{Limitations and conclusions}
The primary limitations of this work are its focus on examples of two-dimensional distributions and a small number of scoring methods; it does not investigate higher dimensions nor attempt to systematically discover and/or evaluate all possible equipoint or equidensity scores. We note that these scores do apply in higher dimensions; however as dimensionality rises, there is risk that the curse of dimensionality will affect equidensity scores as it does, for example, the Jaccard score, leading to artifactually \textit{lower} scores in higher dimensions than one might consider reasonable (in the case of the Jaccard score, because the union grows much faster than the intersection as dimensionality increases). How a one-dimensional version of the Eden score would compare to the KS statistic or a one-dimensional KL score, is an interesting question. Studies of higher dimensions might be considered useful, since non-trivial real datasets are often high-dimensional (e.g., many columns), but using human raters as the gold standard would be complicated by the inability to easily visualize higher dimensions and by potentially important losses if dimensionality reduction is used (e.g. PCA, tSNE, UMAP) \cite{jolliffe_principal_2016, maaten_visualizing_2008, mcinnes_umap_2020}. Fortunately, lower-order relationships often carry a large amount of information about the configuration of complex systems \cite{bialek_rediscovering_2007, schneidman_weak_2006}. For this reason, we expect Eden and other equidensity scores to be useful additions to the generative-modeling toolkit.

\bibliographystyle{unsrt}

\end{document}